\documentclass{article}
\usepackage{spconf,amsmath,graphicx}
\usepackage{amssymb}
\usepackage{multirow}
\usepackage{subcaption}
\usepackage{hyperref}

\usepackage{xcolor}

\usepackage{pifont}% http://ctan.org/pkg/pifont
\usepackage{algorithm}
\usepackage{listings}
\newcommand{\xmark}{\ding{55}}%
\usepackage{etoolbox}

\newcommand{\lnorm}[1]{\frac{#1}{\left\lVert{#1}\right\rVert _2}}

\title{Shifting Focus: From Global Semantics to Local Prominent Features in Swin-Transformer for Knee Osteoarthritis Severity Assessment}

\name{ \begin{tabular}{c}
Aymen Sekhri$^{1*}$\thanks{*Authors with equal contributions.}, Marouane Tliba$^{1*}$, Mohamed Amine Kerkouri$^{1*}$, Yassine Nasser$^{1}$,\\ Aladine Chetouani$^{1}$, Alessandro Bruno$^{2}$, and Rachid Jennane$^{3}$
\end{tabular}
}

\address{$^{1}$Laboratoire PRISME, Université d'Orléans, Orléans, France\\
$^{2}$  IULM Libera Università di Lingue e Comunicazione. 20143 Milan, Italy\\
$^{3}$IDP - UMR CNRS 7013, Université d’Orléans. Orléans, France \\}

\begin{document}

\maketitle

\begin{abstract}
Conventional imaging diagnostics frequently encounter bottlenecks due to manual inspection, which can lead to delays and inconsistencies. Although deep learning offers a pathway to automation and enhanced accuracy, foundational models in computer vision often emphasize global context at the expense of local details, which are vital for medical imaging diagnostics. To address this, we harness the Swin Transformer's capacity to discern extended spatial dependencies within images through the hierarchical framework. Our novel contribution lies in refining local feature representations, orienting them specifically toward the final distribution of the classifier. This method ensures that local features are not only preserved but are also enriched with task-specific information, enhancing their relevance and detail at every hierarchical level. By implementing this strategy, our model demonstrates significant robustness and precision, as evidenced by extensive validation of two established benchmarks for Knee OsteoArthritis (KOA) grade classification. These results highlight our approach's effectiveness and its promising implications for the future of medical imaging diagnostics. Our implementation is available on  \href{https://github.com/mtliba/KOA_NLCS2024}{Github.}
\end{abstract}

\begin{keywords}
Medical Imaging Diagnostics, Representation Learning, Knee OsteoArthritis (KOA) Grade Classification, Task-Relevant Feature Representation
\end{keywords}

\vspace{-3mm}

\section{Introduction}
\label{sec:intro}
\vspace{-3mm}
%Knee OsteoArthritis (KOA) is one of the most prevalent forms of arthritis, primarily affecting the elderly population. It is a degenerative condition that targets the knee joint and usually results from wear and tear, along with a progressive loss of articular cartilage. In some cases, infections can damage the joint cavity, leading to discomforts such as mobility limitations, joint pain, and swelling \cite{tiulpin2020automatic}. Radiographic imaging reveals essential visual markers of KOA, including joint space narrowing, the presence of bone spurs, and sclerosis. Despite the availability of various diagnostic tools such as magnetic resonance imaging, computed tomography, and ultrasound, radiography remains the preferred initial method due to its accessibility and cost-effectiveness.To quantify the severity of KOA, clinicians commonly rely on the Kellgren and Lawrence (KL) grading system, which offers a five-stage classification ranging from KL grade 0 (Normal) to KL grade 4 (Severe) \cite{kohn2016classifications,kellgren1957radiological}. However, it's important to note that KOA tends to develop progressively, introducing subjectivity into its classification into distinctive grades. This subjectivity poses challenges in automating KOA diagnosis, especially given the striking similarity between X-ray images, making accurate diagnosis a more intricate task.

Knee OsteoArthritis (KOA) stands as a prevalent arthritis form, majorly afflicting the elderly, characterized by the degenerative wear and tear of knee joint cartilage. Occasionally, joint infections exacerbate this degeneration, manifesting as mobility constraints, pain, and swelling. Although various imaging modalities exist, radiography is preferred for initial KOA assessment due to its cost-effectiveness and widespread availability, revealing critical indicators like joint space narrowing and bone spur formation.
Clinicians traditionally employ the Kellgren and Lawrence (KL) grading system \cite{kellgren1957radiological} to evaluate KOA severity, spanning five stages from normal (KL-0) to severe (KL-4). Despite its utility, KOA's gradual progression introduces a degree of subjectivity in these classifications, complicating the automation of diagnosis. The nuanced distinctions necessary for accurate KL grade determination from X-ray images underline the inherent challenges in automating KOA diagnosis, necessitating advanced solutions to enhance diagnostic precision and reliability.

Despite the advancements in deep learning models for image recognition, a persistent limitation exists in their ability to capture detailed local features and construct hierarchical representations concurrently, as the opted training strategies for this purpose push the network to be invariant to local feature transformation. In contrast, this limitation is particularly pronounced in medical imaging, where the nuanced differentiation of local textural features within a joint is crucial. For KOA diagnosis, the ability to discern subtle variations in joint texture while contextualizing the overall joint form is paramount \cite{KOA_texture}. Traditional deep-learning approaches and training strategies tend to excel in identifying broad patterns but often falter at isolating minute yet diagnostically critical details. This inadequacy is especially detrimental in the context of KOA, where the early detection of subtle textural changes can significantly influence treatment outcomes \cite{KOA_texture}. As such, there is a pressing need for a training strategy that can adeptly navigate the intricacies of medical images, capturing the fine-grained details necessary for accurate diagnosis without compromising on the broader contextual understanding. 

Addressing this gap, our research makes two fundamental contributions:  
\begin{itemize}
\setlength\itemsep{1em}
\item  We leverage the Swin Transformer architecture's ability to capture localized features at each layer with a long range of features space dependencies, combining this information with the global representation ensures a comprehensive representation that integrates both relevant local features and global information. 

\item  We refine the final representation by employing a regularization term: we optimize the alignment between local features and the network's decision-making features distribution layer using a Negative Cosine Similarity Loss (NCSL). This technique works as a regularization term to the hierarchical representation abstraction learning process. ensuring critical local details are preserved and emphasized during the classification process.
\end{itemize}

This approach ensures that while the network maintains its robustness to global context and transformations, it does not sacrifice the critical local information essential for precise medical diagnostics, thereby aligning the model's learning process with the unique demands of medical imaging in KOA.

\section{Related Works}
\vspace{-3mm}
In Knee Osteoarthritis (KOA) diagnosis, deep learning (DL) methodologies have significantly advanced, leveraging rich datasets from the Osteoarthritis Initiative (OAI) \cite{OAI} and the Multicenter Osteoarthritis Study (MOST) \cite{MOST}. Various approaches have sought to harness the potential of DL in discerning KOA severity and progression.
Noteworthy endeavors include Antony et al.\cite{antony2017automatic}, who utilized Convolutional Neural Networks (CNNs) for a two-stage assessment involving knee joint localization followed by classification, innovatively combining classification and regression losses for enhanced accuracy. Similarly, Tiulpin et al. \cite{tiulpin2020automatic} employed a Siamese CNN structure to estimate KOA KL grades, amalgamating multiple model outputs for robust predictions. Chen et al. \cite{chen2019fully} introduced an ordinal loss-based strategy to optimize CNNs for KOA grading, leveraging the KL scale's inherent ordinality. Nasser et al. \cite{nasser2020discriminative} further contributed by developing a Discriminative Regularized Auto-Encoder (DRAE) targeting early KOA detection, emphasizing feature distinction via a discriminative penalty.
Emerging research by Wang et al. \cite{wang2021automatic} explored Vision Transformer architectures, complemented by novel data augmentation methods and a hybrid loss function to address early KOA identification and tackle data drift challenges in multi-center studies, a topic we have previously addressed to enhance cross-dataset model efficacy.
Nasser et al. \cite{nasser2022discriminative} devised a deep neural network that synergizes shape and texture analysis.

\section{Proposed Method}
\label{sec:proposed_method}
\begin{figure}[]
    \centering
    \includegraphics[scale = 0.41]{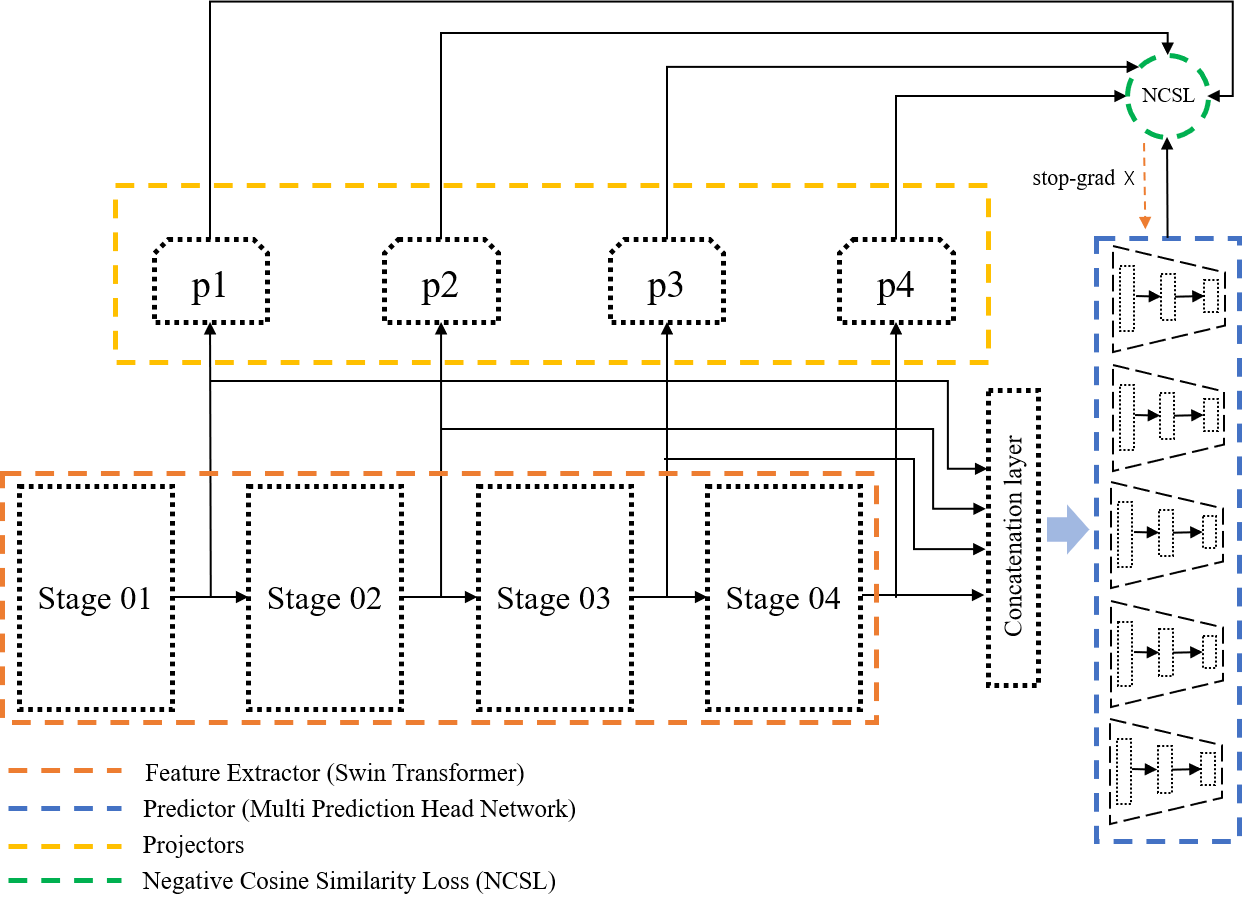}
    \setlength{\belowcaptionskip}{-3mm}
    \caption{Overview of the Proposed KOA Diagnostic Model. This figure outlines our advanced model, highlighting the Swin Transformer for hierarchical feature extraction, multiple prediction heads for detailed KL grade classification, skip connections for effective feature flow, and Negative Cosine Similarity Loss for feature optimization. Together, these elements illustrate our novel approach to balancing local detail recognition with global feature abstraction for improved KOA diagnostic accuracy.}
    \label{fig:arch}
    \vspace{-2mm}
\end{figure}

\vspace{-3mm}
Our proposed methodology integrates four key components: (1) A Swin Transformer is utilized as the primary feature extractor, capitalizing on its efficacy in hierarchical feature representation. (2) A multi-prediction head network is deployed for classification, ensuring robust and accurate categorization. (3) Skip connections are integrated within the architecture to facilitate enhanced information flow and feature retention. (4) The adoption of Negative Cosine Similarity Loss (NCSL) across the transformer stages serves as a regularization term, optimizing local feature refinement and aligning it with the global representation, thus enhancing model performance. For the architectural visualization, refer to Figure\ref{fig:arch}.

\subsection{Swin Transformer}
\label{subsec:swin}
The Swin Transformer, delineated in \cite{liu2021swin}, is a pivotal advancement in adapting transformers for computer vision, featuring hierarchical feature maps and an optimized Multi-head Self Attention (MSA) framework to improve efficiency and effectiveness. Initially, the model partitions input images of size $H \times W \times 3$ into $4 \times 4$ patches, which are then embedded and processed through the model's stages. The initial stage involves a linear embedding, followed by two Swin Transformer blocks using Window Multi-head Self-Attention (W-MSA) and Shifted Window Multi-head Self-Attention (SW-MSA). Subsequent stages merge and transform these patches, diminishing their quantity while increasing dimensionality through additional Swin Transformer blocks. By its final stage, the model refines the patches to a significantly reduced count with augmented feature representation, underscoring the Swin Transformer's utility in computer vision tasks.

\subsection{Multi-Prediction Head Network}
\label{subsec:mphn}
The Multi-Prediction Head Network (MPHN), integral to our methodology and elaborated upon in prior work \cite{CBMI_KOA}, deconstructs the multi-class task into simpler binary classifications. It features a series of Multilayer Perceptrons (MLPs), each with three layers, designed to identify individual KL grades, thereby encompassing five distinct MLPs tailored for specific grade classifications.

\subsection{Skip Connection}
\label{subsec:skip_conn}
Medical imaging diagnostics emphasize localized features, contrasting with the broader focus typical of natural image analysis. Utilizing the Swin Transformer's hierarchical feature mapping, as expounded in Section \ref{subsec:swin}, we derive and normalize stage-specific outputs. These, coupled with an average pooling step, culminate in refined local feature representations. By amalgamating these with the comprehensive output from the final stage, our model achieves a synergistic a balanced consideration of both detailed local attributes and global semantic context.

However, the amalgamation of the features representation is not valid without a distillation of the representation for the following considerations, first, the local features space may contain noises that shift the deep semantic representation, second, the two representations (local features, and global features) should be in the same features space dimension.

\subsection{Optimization of Hierarchical Local Feature Representation}

\label{subsec:enhance_local}
Our methodology focuses on enhancing the hierarchical representation of local features, we took insight from advances in Knowledge 
distillation \cite{zhu2018knowledge} and self-supervised learning \cite{grill2020bootstrap,chen2021exploring} \cite{mtliba} from features similarities. We formulate our contribution as a regularization term that helps the network to keep the prominent local features within Swin transformer, denoted as a learnable function $f_{\theta}$, where $\theta$ represents the model parameters. For each stage $s$ of the transformer, we obtain an output $O_s = f_{\theta}(I_s)$, where $I_s$ represents the input to that stage.
To refine these outputs further and align their representation spaces, we introduce projection heads $P_{\phi_s}$ for each stage $s$, parameterized by $\phi_s$. The projected representation for each stage is given by $P_s = P_{\phi_s}(O_s)$.
The concatenated output $C$ of these projected representations from all stages forms the enhanced feature representation and is given by:

\begin{equation}
C = \bigoplus_{s=1}^{S} P_s
\end{equation}

where $\bigoplus$ denotes the concatenation operation, and $S$ is the total number of stages.

For classification, building upon the preliminary context discussed in Section \ref{subsec:mphn}, we instantiate separate classifier networks, denoted as $C_{\psi_k}$ for each class $k$, where each network is parameterized by $\psi_k$. This architecture enables tailored processing for each specific class, ensuring independent and specialized distillation of the relevant KL grade representation  during classification. Each classifier network, conceptualized to encapsulate the feature' space distribution pertinent to its designated class, functions as a sophisticated decision maker in our framework.

Formally, the operation of the classifier for each class $k$ can be delineated as follows: The decision maker network processes the concatenated feature representation $C$, yielding an intermediate output $D_k = C_{\psi_k}(C)$. To transition from this multidimensional output to a scalar prediction value, we introduce a learnable parameter vector $\omega_k$ for each class $k$, facilitating the necessary dimensionality reduction. Consequently, the predictive likelihood $y_k$ for each class is computed via a sigmoid activation function $\sigma$, applied to the dot product of $D_k$ and $\omega_k$:

\begin{equation}
    y_k = \sigma(\omega_k^\top D_k)
\end{equation}

where $\sigma(\cdot)$ denotes the sigmoid function, ensuring that the final output $y_k$ resides within the interval $[0, 1]$, representing the probability or likelihood of class $k$ being the correct classification.

The empirical risk minimization for each class $k$ is conducted using Binary Cross-Entropy (BCE), formulated as:

\begin{equation}
\mathcal{L}_{BCE_k} = -\sum_{n} [y_n^k \log(\hat{y}_n^k) + (1 - y_n^k) \log(1 - \hat{y}_n^k)]
\end{equation}

where $y_n^k$ and $\hat{y}_n^k$ are the true and predicted labels for the $n$-th sample and class $k$, respectively.

In parallel, we aim to minimize the Negative Cosine Similarity Loss (NCSL) between the projected features $P_s$ and the features distribution from the classifier, enhancing the feature space's relevance. The NCSL inspired from \cite{chen2021exploring} is defined as:

\begin{equation}
\mathcal{L}_{NCSL} = -\frac{1}{S} \sum_{s=1}^{S}  \lnorm{P_s}{\cdot}\lnorm{\texttt{sg}(D)}
\end{equation}

%where $D$ denotes the feature distribution from the set of $k$ decision makers, and $sg$ refers to the stop gradient $P_s$, we opted to use only one flow as our goal is to add regularization term that alligne the local features towards the featuers distribution of the clssifier in order to .

where $D$ represents the aggregated feature distribution across $k$ decision-makers. The function $\texttt{sg}(\cdot)$ denotes the stop-gradient operation applied to $D$, essential for detaching the main branch during optimization. This selection is intentional, focusing on a singular optimized pathway to reinforce the regularization effect. The primary objective here is to meticulously align the local feature representations $P_s$ with the global classifier-driven feature landscape $D$ as ground truth target, fostering a synergistic enhancement in feature discernment and classification efficacy. Our implementation seeks to deeply integrate these local representations with the classifier's insights, ensuring a robust and contextually aware feature synergy for improved model interpretability and performance.

The total loss $\mathcal{L}$ combines the BCE loss for each class and the NCSL, providing a comprehensive learning signal:

\begin{equation}
\mathcal{L} = \sum_{k=1}^{K} \mathcal{L}_{BCE_k} + \lambda \mathcal{L}_{NCSL}
\end{equation}

where $K$ is the total number of classes and $\lambda$ is a  coefficient balancing the two loss components. This loss function orchestrates the overall learning process, emphasizing both global risk minimization and the fidelity of local feature representations.

\section{result analysis}
\label{sec:pagestyle}
% \subsection{Quantitative Evaluation}
% \label{subsec:quant}
\vspace{-3mm}

\subsection{Dataset}
\label{subsec:dataset}

In this study, we employed two commonly used databases, the OsteoArthrit Initiative (OAI) \cite{OAI} and the Multicenter Osteoarthritis Study (MOST) \cite{MOST}. We adopted the identical data preprocessing approach as described in our earlier publication \cite{CBMI_KOA}. However, a noteworthy departure in our methodology was the use of the MOST dataset for training our feature extractor. Subsequently, we employed the pre-trained weights from this feature extractor as an initialization step for training on the OAI dataset. 

% \textcolor{blue}{, thereby applying our two primary contributions}
% \textcolor{blue}{Added a table of the data distribution used in this study.}

% \textcolor{red}{Is it a binary or multiclassification task of the grades? It is not clear to me.}

% \subsection{Ablation Study}
% \label{subsec:ablation}
% For the sake of our ablation study to show the effectiveness of our proposed approach, we use three different decision makers. Namely the Single Prediction Head Network (SPHN) in which we used a single MLP for predicting all KL grades in multi class classification manner, the second by using the Multi Prediction Head Network (MPHN) as illustrated in section \ref{subsec:mphn}, and finally we used an MLPRegressor, by considering the task as a regression problem.

\begin{table}[ht]
    \centering
    \small
    \begin{tabular}{c|c|c|c|c|c|c|c}
        \hline
        \setlength{\tabcolsep}{2pt}
        \multirow{2}{*}{Setups} & \multicolumn{4}{c|}{Method} & \multicolumn{3}{c}{Metrics} \\
        \cline{2-8}
         & A & B & C & D & ACC (\%) & B-ACC (\%) & F1-Score \\
        \hline
        1 & \checkmark & \xmark & \xmark & \xmark & 70.89 & 69.07 & 0.698 \\
        2 & \checkmark & \xmark & \xmark & \checkmark & 71.68 & 69.13 & 0.687 \\
        \hline
        3 & \xmark & \checkmark & \xmark & \xmark & 71.44 & 67.46 & 0.690 \\
        \textbf{4} & \xmark & \checkmark & \xmark & \checkmark & \textbf{72.40} & \textbf{71.11} & \textbf{0.704} \\
        \hline
        5 & \xmark & \xmark & \checkmark & \xmark & 66.55 & 54.05 & 0.528 \\
        6 & \xmark & \xmark & \checkmark & \checkmark & 67.63 & 54.29 & 0.534 \\
        \hline
    \end{tabular}
    \caption{Results of the ablation study using the OAI test-set. A: SPHN, B: MPHN, C: MLPReg, D: NCSL}
    \label{tab:ablation}
    \vspace{-9mm}
\end{table}

\subsection{Ablation Study}
\label{subsec:ablation_summary}

Our ablation study evaluates the efficacy of different decision frameworks within our model, contrasting three configurations: a Single Prediction Head Network (SPHN) that employs a universal MLP across all KL grades, a Multi Prediction Head Network (MPHN) as elaborated in Section \ref{subsec:mphn}, and an MLPRegressor that conceptualizes the task in a regression paradigm.

Key observations from our analysis, delineated in Table \ref{tab:ablation}, are summarized as follows:
 \textbf{(1)SPHN Analysis:} In Setup 1, we employed SPHN with integrated feature concatenation from the Swin Transformer, establishing our baseline. Enhancement was observed in Setup 2 where the NCSL application improved ACC from 70.89\% to 71.68\% and yielded a modest uplift in F1-Score.
\textbf{(2)MPHN Evaluation:} Advancing to Setups 3 and 4 with the MPHN, marked improvements were noted. Significantly, Setup 4, which synergizes MPHN with NCSL, recorded the highest metrics: ACC (72.40\%), B-ACC (71.11\%), and F1-Score (0.704), attesting to the MPHN's capacity to distinctively refine and exploit grade-specific features.
\textbf{(3)MLPRegressor Results:} The last frameworks, Setups 5 and 6, implementing an MLPRegressor, demonstrated marginal enhancements, peaking at an ACC of 67.63\%, suggesting a constrained utility of the regression format for this classification context.
To sum up , these findings substantiate the integration of MPHN and NCSL as the most efficacious approach.

\subsection{Qualitative Analysis}

\begin{figure}[!ht]

\includegraphics[width=\columnwidth]{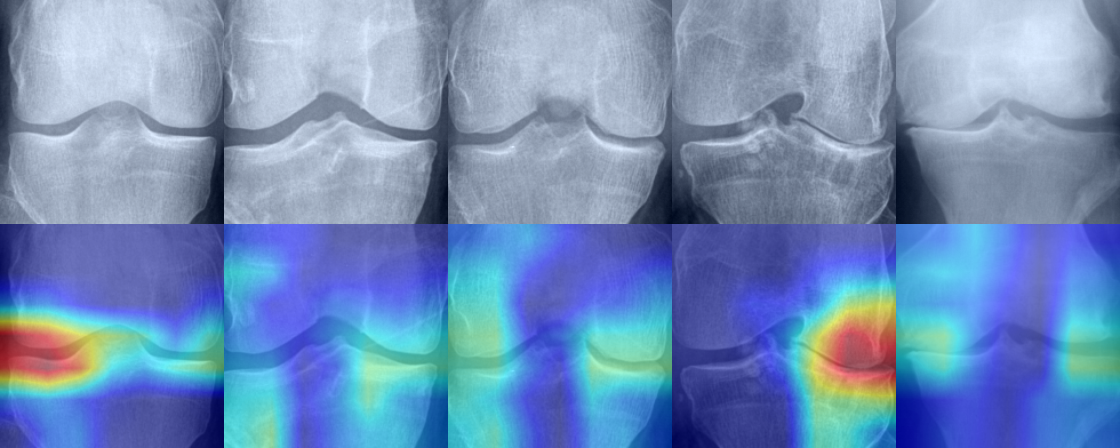} \caption{ \label{figure:gradCAMs}
\cite{selvaraju2017grad} demonstrate our model's ability to discern progression in KOA severity from KL grade 0 (healthy) to KL grade 4 (most severe). These visualizations underscore the model's focus on both local and global features, adjusting according to severity. They confirm our model's effective use of various hierarchical feature levels without central region bias, relaying on both joint' local textures and global form}
\vspace{-3mm}
\end{figure}

We used GradCAM \cite{selvaraju2017grad} as a tool to visualize the activations of the final layer of the feature extractor. Within our analysis, we carefully selected representative samples from each grade of knee OA (KL-0 to KL-4, ordered left to right). Visualizations are presented in Figure \ref{figure:gradCAMs}. 
As can be seen this Figure reveals that the model adeptly identifies crucial features such as osteophytes, joint space narrowing, and sclerosis. These elements hold significant clinical importance in assessing the severity of knee OA \cite{lespasio2017knee}. %This underscores the notion that our model's classifications are rooted in relevant regions of interest commonly employed in clinical diagnosis, rather than being influenced by irrelevant features.

\subsection{Comparison to State-Of-The-Art (SOTA)}
\label{subsec:sota}
Table \ref{tab:SOTAComp} presents a comparison of the  obtained results to SOTA methods.

\begin{table}[htbp!]
    \small
    \begin{center}
    \begin{tabular}{ l c c  }
    \hline
    \textbf{Model} & \textbf{Accuracy (\%) $\uparrow$ } & \textbf{F1-Score $\uparrow$ }  \\ 
    \hline
    Antony et al. 2016 \cite{antony2016quantifying} & 53.40 & 0.43 \\ \hline
    Antony et al. 2017 \cite{antony2017automatic} & 63.60 & 0.59 \\ \hline
    Tiulpin et al. 2018 \cite{tiulpin2018automatic} & 66.71 & - \\ \hline
    Chen et al. (Vgg19) 2019 \cite{chen2019fully} & 69.60 & - \\ \hline
    Chen et al. (ResNet50) 2019 \cite{chen2019fully} & 66.20 & - \\ \hline
    Chen et al. (ResNet101) 2019 \cite{chen2019fully} & 65.50 & - \\ \hline
    Wang et al. 2021 \cite{wang2021automatic} & 69.18 & - \\ \hline
    Sekhri et al. 2023 \cite{CBMI_KOA} & 70.17 & 0.67 \\ \hline
    \textbf{Ours - Setup 4} & \textbf{72.40} & \textbf{0.70} \\ \hline
    \end{tabular}
    \caption{\label{tab:SOTAComp} State of the art Comparison on OAI Test set}
    \vspace{-7mm}
    \end{center}
\end{table}

Antony et al. \cite{antony2016quantifying} and \cite{antony2017automatic} reported their diagnostic accuracies as 53.40\% and 63.60\%, respectively, accompanied by F1-scores of 0.43 and 0.59. Meanwhile, Chen et al. \cite{chen2019fully} employed an ordinal loss strategy with various deep learning architectures. They achieved accuracies of 69.60\%, 66.20\%, and 65.50\% using VGG19, ResNet50, and ResNet101, respectively. However, the corresponding F1 scores were not provided in their study. Tiulpin et al. \cite{tiulpin2018automatic} harnessed a Siamese network, achieving an ACC of 66.71\%. Furthermore, Wang et al. \cite{wang2021automatic} demonstrated an accuracy of 69.18\%. Our proposed method surpasses all aforementioned previous methods, with an ACC of $70.17\%$ and a F1-score of $0.67$. 

Our contributions, specifically setup 4 (Table \ref{tab:ablation}), stands out as the best approach, surpassing all previous works. Notably, it achieves an enhanced accuracy of $72.40\%$ and an improved F1-Score of $0.70$. These findings reflect the ongoing refinement of our methodology by preserving the learned local features as well as minimizing the NCSL between the output features at each stage and the last output hidden layers inside the MLPs helped to bolster the overall Accuracy.

\section{conclusion}
\label{sec:typestyle}
\vspace{-3mm}
%In conclusion, our paper introduces an innovative approach employing the Swin Transformer network for predicting Knee OA severity from radiographic images. Our work demonstrates that this method establishes a new benchmark by achieving state-of-the-art performance on the OAI test dataset, surpassing significantly existing methodologies.
%Our study underscores the efficacy of the Swin Transformer network in extracting pertinent knee OA information. Thereby enabling the detection of a wide range of disease symptoms. Furthermore, the integration of learned local features at each Swin Transformer stage with global features, along with the implementation of a multi-prediction head network, contributes substantially to the model's accuracy and minimizes feature similarity between adjacent grades.

%Maximizing the shared information between features extracted from each stage and the final hidden layer within the Multi-Prediction Head Network significantly enhances performance, underscoring the significance of our proposed approach. 

In conclusion, our study presents a comprehensive evaluation of an advanced model tailored for the task of KOA grade classification.  Throughout an extensive experiment, we have demonstrated the superior performance of our proposed approach, particularly when employing the MPHN in conjunction with NCSL as regularization for feature refinement. The results indicate our model accuracy and reliability of KOA diagnostics.  Furthermore, our proposed novel approach underscores the significance of adopting a specialized, hierarchical feature processing strategy. This research opens a significant path forward for the application of deep learning for medical imaging, paving the way for more precise and dependable automatic diagnostic tools in the field of computational radiology.
\small
\bibliographystyle{IEEEbib}
\bibliography{strings,refs}

\end{document}